\documentclass{article}
\usepackage{ijcai18}

\usepackage{times}
\usepackage{xcolor}
\usepackage{soul}
\usepackage[utf8]{inputenc}
\usepackage[small]{caption}
\usepackage{url}  
\usepackage{graphicx}
\usepackage{subfigure}
\usepackage{amsmath}
\usepackage{amsfonts}
\usepackage{multirow}
\usepackage{arydshln}
\usepackage{balance}

\title{Neural Machine Translation with Key-Value Memory-Augmented Attention}

\author{
Fandong Meng,
Zhaopeng Tu,
Yong Cheng,
Haiyang Wu,
Junjie Zhai,
Yuekui Yang,
Di Wang
\\
Tencent AI Lab \\
\{fandongmeng,zptu,yongcheng,gavinwu,jasonzhai,yuekuiyang,diwang\}@tencent.com
}

\begin{document}

\maketitle

\begin{abstract}
Although attention-based Neural Machine Translation (NMT) has achieved remarkable progress in recent years, it still suffers from issues of repeating and dropping translations. To alleviate these issues, we propose a novel \emph{key-value memory-augmented attention} model for NMT, called \textsc{KVMemAtt}. Specifically, we maintain a timely updated key-memory to keep track of attention history and a fixed value-memory to store the representation of source sentence throughout the whole translation process. Via nontrivial transformations and iterative interactions between the two memories, the decoder focuses on more appropriate source word(s) for predicting the next target word at each decoding step, therefore can improve the adequacy of translations. 
Experimental results on Chinese$\Rightarrow$English and WMT17 German$\Leftrightarrow$English translation tasks demonstrate the superiority of the proposed model.
\end{abstract}

\section{Introduction}
The past several years have witnessed promising progress in Neural Machine Translation (NMT)~\cite{ChoEMNLP,googleS2S},
in which attention model plays an increasingly important role~\cite{cho,luongEMNLP2015,VaswaniEtal2017}. Conventional attention-based NMT encodes the source sentence as a sequence of vectors after bi-directional recurrent neural networks (RNN)~\cite{schuster1997bidirectional}, and then generates a variable-length target sentence with another RNN and an attention mechanism. The attention mechanism plays a crucial role in NMT, as it shows which source word(s) the decoder should focus on in order to predict the next target word.
However,  there is no mechanism to effectively keep track of attention history in conventional attention-based NMT. This may let the decoder tend to ignore past attention information, and lead to the issues of repeating or dropping translations~\cite{Tu2016}. For example, conventional attention-based NMT may repeatedly translate some source words while mistakenly ignore other words.

A number of recent efforts have explored ways to alleviate the inadequate translation problem. For example,~\citeauthor{Tu2016}~\shortcite{Tu2016} employ coverage vector as a lexical-level indicator to indicate whether a source word is translated or not.~\citeauthor{mengEtAlCOLING2016}~\shortcite{mengEtAlCOLING2016} and~\citeauthor{Zheng:2018:TACL}~\shortcite{Zheng:2018:TACL} take the idea one step further, and directly model translated and untranslated source contents by operating on the attention context (i.e., the partial source content being translated) instead of on the attention probability (i.e., the chance of the corresponding source word is translated). Specifically, ~\citeauthor{mengEtAlCOLING2016}~\shortcite{mengEtAlCOLING2016} capture translation status with an interactive attention augmented with a NTM~\cite{ntmgraves2014neural}  memory.~\citeauthor{Zheng:2018:TACL}~\shortcite{Zheng:2018:TACL} separate the modeling of translated (\textsc{Past}) and untranslated (\textsc{Future}) source content from decoder states by introducing two additional decoder adaptive layers. 

\citeauthor{mengEtAlCOLING2016}~\shortcite{mengEtAlCOLING2016} propose a generic framework of memory-augmented attention, which is independent from the specific architectures of the NMT models.
However, the original mechanism takes only a single memory to both represent the source sentence and track attention history. Such overloaded usage of memory representations makes training the model difficult~\cite{Rocktaschel:2017:ICLR}.
In contrast,~\citeauthor{Zheng:2018:TACL}~\shortcite{Zheng:2018:TACL} try to ease the difficulty of representation learning by separating the \textsc{Past} and \textsc{Future} functions from the decoder states. However, it is designed specifically for the precise architecture of NMT models.

In this work, we combine the advantages of both models by leveraging the generic memory-augmented attention framework, while easing the memory training by maintaining separate representations for the expected two functions. Partially inspired by~\cite{millerEMNLP2016}, we split the memory into two parts: a {\em dynamical key-memory} along with the update-chain of the decoder state to keep track of attention history, and a {\em fixed value-memory} to store the representation of source sentence throughout the whole translation process.
In each decoding step, we conduct multi-rounds of memory operations repeatedly layer by layer, which may let the decoder have a chance of re-attention by considering the ``intermediate'' attention results achieved in early stages.
This structure allows the model to leverage possibly complex transformations and interactions between 1) the key-value memory pair in the same layer, as well as 2) the key (and value) memory across different layers.

Experimental results on Chinese$\Rightarrow$English translation task show that attention model augmented with a single-layer key-value memory improves both translation and attention performances over not only a standard attention model, but also over the existing NTM-augmented attention model~\cite{mengEtAlCOLING2016}. Its multi-layer counterpart further improves model performances consistently. We also validate our model on bidirectional German$\Leftrightarrow$English translation tasks, which demonstrates the effectiveness and generalizability of our approach.

\section{Background} \label{back}

\paragraph{Attention-based NMT}
Given a source sentence $\mathbf{x} \hspace{-3pt}=\hspace{-3pt} \{x_1,x_2,\cdots, x_n\}$ and a target sentence $\mathbf{y} \hspace{-3pt}=\hspace{-3pt} \{y_1,y_2,\cdots, y_m\}$, NMT models the translation probability word by word:
\begin{eqnarray}
p(\mathbf{y}|\mathbf{x}) &=& \prod_{t=1}^{m}{P(y_{t}|\mathbf{y_{<t}}, \mathbf{x}; \theta)} \nonumber \\
                         &=& \prod_{t=1}^{m}{softmax(f(\mathbf{c}_t, y_{t-1}, \mathbf{s}_t))} \label{predict}
\end{eqnarray}
where $f(\cdot)$ is a non-linear function, and $\mathbf{s}_t$ is the hidden state of decoder RNN at time step $t$:
\begin{eqnarray}
\mathbf{s}_t = g(\mathbf{s}_{t-1}, y_{t-1}, \mathbf{c}_t) \label{dec_state_update}
\end{eqnarray}
$\mathbf{c}_t$ is a distinct source representation for time $t$, calculated as a weighted sum of the source annotations:
\begin{eqnarray}
\mathbf{c}_t = \sum_{j=1}^{n}{a_{t,j} \mathbf{h}_j} \label{attention_softmax}
\end{eqnarray}
where $\mathbf{h}_j$ is the encoder annotation of the source word $x_j$,
and the weight $a_{t,j}$ is computed by
\begin{eqnarray}
a_{t,j} = \frac{exp(e_{t,j})}{\sum_{k=1}^{N}{exp{(e_{t,k})}}} \label{match_score}
\end{eqnarray}
where $e_{t,j}=\mathbf{v}_a^Ttanh(\mathbf{W}_a \tilde{s}_{t-1} + \mathbf{U}_a  \mathbf{h}_j)$ scores how much $\tilde{\mathbf{s}}_{t-1}$ attends to $\mathbf{h}_j$, where $\tilde{\mathbf{s}}_{t-1}=g(\mathbf{s}_{t-1}, y_{t-1})$ is an intermediate state tailored for computing the attention score with the information of $y_{t-1}$. 

The training objective is to maximize the log-likelihood of the training instances ($\mathbf{x}^s$, $\mathbf{y}^s$):
\begin{eqnarray}
\mathbf{\theta^{*}} =  \arg\max_{\theta}\sum_{s=1}^{S} {\log p(\mathbf{y}^{s}|\mathbf{x}^{s})} \label{nmt_obj}
\end{eqnarray}

\begin{figure}[t!]
\begin{center}
      \includegraphics[width=0.44\textwidth]{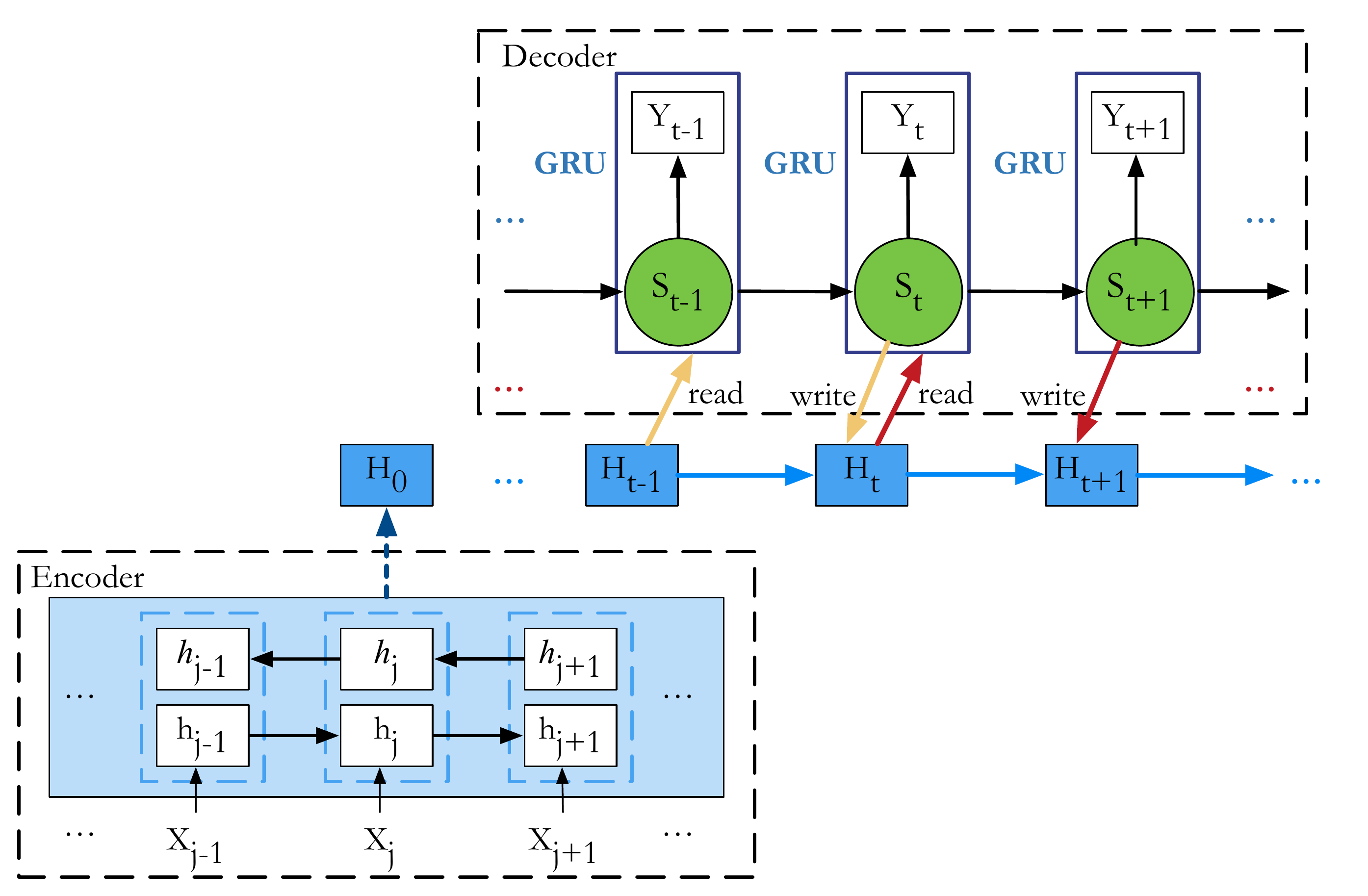} 
       \caption{Illustration for the NTM-augmented attention. The yellow and red arrows indicate interactive read-write operations.} \label{f:inter} \vspace{-10pt}
  \end{center} \vspace{-5pt}
\end{figure}

\begin{figure*}[t!]
\begin{center}
      \includegraphics[width=0.8\textwidth]{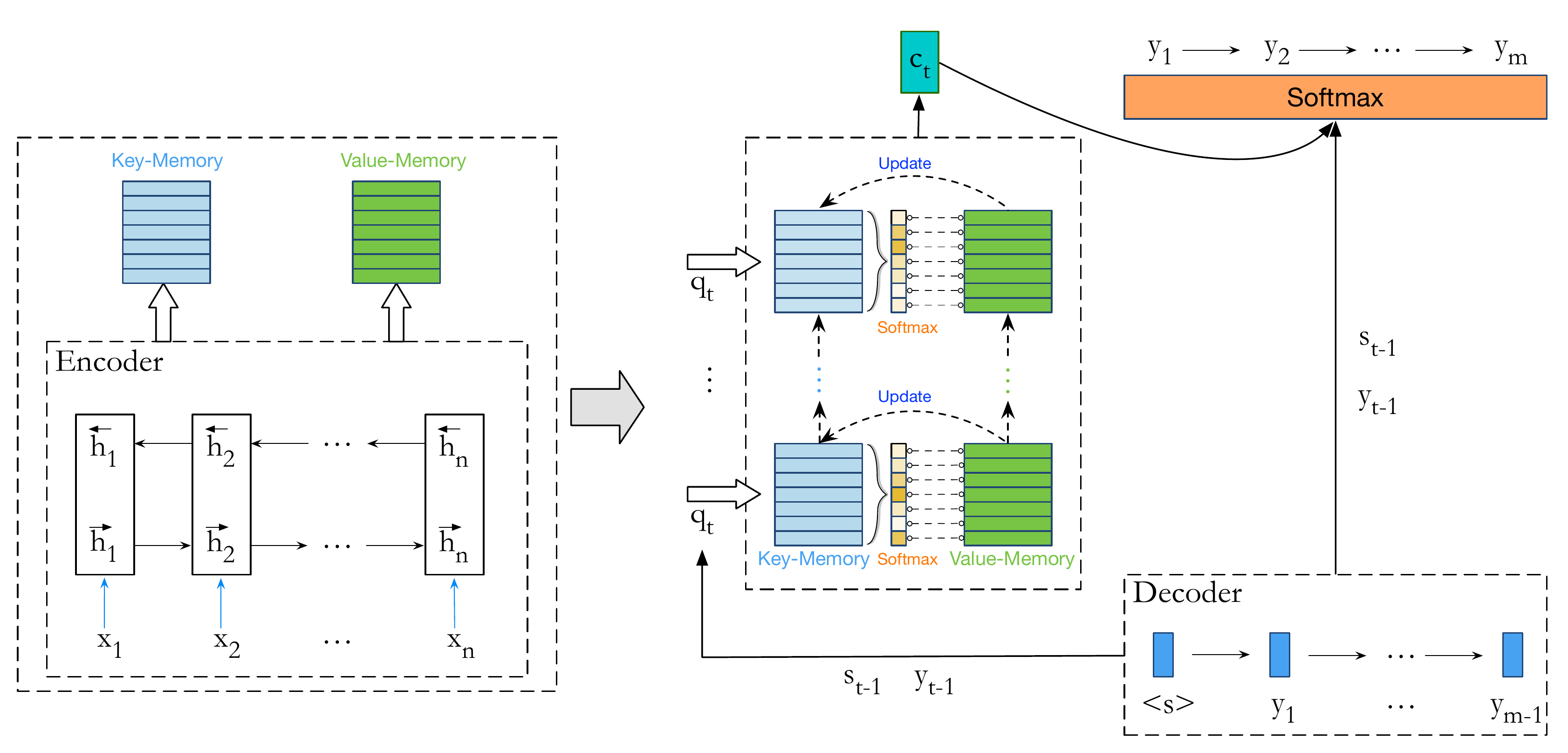}  \vspace{-5pt}
       \caption{The architecture of \textsc{KVMemAtt}-based NMT. The encoder (on the left) and the decoder (on the right) are connected by the \textsc{KVMemAtt} (in the middle). In particular, we show the \textsc{KVMemAtt} at decoding time step $t$ with multi-rounds of memory access.} \label{f:mea_nmt}  \vspace{-10pt}
 \end{center}
\end{figure*}

\vspace{-10pt}
\paragraph{Memory-Augmented Attention}
\citeauthor{mengEtAlCOLING2016}~\shortcite{mengEtAlCOLING2016} propose to augment the attention model with a memory in the form of NTM, which aims at tracking the attention history during decoding process, as shown in Figure~\ref{f:inter}. At each decoding step, the NTM employs a network-based reader to read from the encoder annotations and output a distinct memory representation, which is subsequently used to update the decoder state. After predicting the target word, the updated decoder state is written back to the memory, which is controlled by a network-based writer. As seen, the interactive read-write operations can timely update the representation of source sentence along with the update-chain of the decoder state and let the decoder keep track of the attention history.

However, this mechanism takes only a single memory to both represent the source sentence and track attention history. Such overloaded usage of memory representations makes training the model difficult~\cite{Rocktaschel:2017:ICLR}.

\section{Model Description}

Figure~\ref{f:mea_nmt} shows the architecture of the proposed {\bf K}ey-{\bf V}alue {\bf Mem}ory-augmented {\bf Att}ention model (\textsc{KVMemAtt}). 
It consists of three components: 1) an encoder (on the left), which encodes the entire source sentence and outputs its annotations as the initialization of the \textsc{Key-Memory} and the \textsc{Value-Memory}; 2) the key-value memory-augmented attention model (in the middle), which generates the context representation of source sentence appropriate for predicting the next target word with iterative memory access operations conducted on the \textsc{Key-Memory} and the \textsc{Value-Memory}; and 3) the decoder (on the right), which predicts the next target word step by step.

Specifically, the \textsc{Key-Memory} and \textsc{Value-Memory} consists of $n$ slots, which are initialized with the annotations of the source sentence $\{\mathbf{h}_1, \mathbf{h}_2, \dots, \mathbf{h}_n\}$.
\textsc{KVMemAtt}-based NMT maintains these two memories throughout the whole decoding process, with the \textsc{Key-Memory} keeping updated to track the attention history, and the \textsc{Value-Memory} keeping fixed to store the representation of source sentence. 
For example, the $j$-th slot ($v_j$) in \textsc{Value-Memory} stores the representation of the $j$-th source word (fixed after generated), and the $j$-th slot ($k_j$) in \textsc{Key-Memory} stores the attention (or translation) status (updated as translation goes) corresponding to the $j$-th source word. 
At step $t$, the decoder state $\mathbf{s}_{t-1}$ first meets the previous prediction $y_{t-1}$ to form a ``query" state $\mathbf{q}_{t}$, which can be calculated as follows
\begin{eqnarray}
\mathbf{q}_{t}=\mathbf{GRU}(\mathbf{s}_{t-1}, \mathbf{e}_{y_{t-1}})
\end{eqnarray}
where $\mathbf{e}_{y_{t-1}}$ is the word-embedding of the previous word $y_{t-1}$. The decoder uses the ``query" state $\mathbf{q}_{t}$ to \emph{address} from the \textsc{Key-Memory} looking for an accurate attention vector $\mathbf{a}_t$, and \emph{reads} from the \textsc{Value-Memory} with the guidance of $\mathbf{a}_t$ to generate the source context representation $\mathbf{c}_t$. After that the \textsc{Key-Memory} is \emph{updated} with $\mathbf{q}_t$ and $\mathbf{c}_t$. 

The memory access (i.e. \emph{address}, \emph{read} and \emph{update}) in one decoding step can be conducted repeatedly, which may let the decoder have a chance of re-attention (with new information added) before making the final prediction.
Suppose there are $R$ rounds of memory access in each decoding step. 
The detailed operations from round $r$-$1$ to round $r$ are as follows:

First, we use the ``query" state $\mathbf{q}_{t}$ to address from $\mathbf{K}_{(t)}^{(r-1)}$ to generate the ``intermediate" attention vector $\tilde{\mathbf{a}}_{t}^{r}$
\begin{eqnarray}
\tilde{\mathbf{a}}_{t}^{r}=\mathbf{Address}(\mathbf{q}_{t}, \mathbf{K}_{(t)}^{(r-1)}) \label{address}
\end{eqnarray}
which is subsequently used as the guidance for reading from the value memory $\mathbf{V}$ to get the ``intermediate" context representation $\tilde{\mathbf{c}}_{t}^{r}$ of source sentence
\begin{eqnarray}
\tilde{\mathbf{c}}_{t}^{r}=\mathbf{Read}(\tilde{\mathbf{a}}_{t}^{r}, \mathbf{V}) \label{read}
\end{eqnarray}
which works together with the ``query" state $\mathbf{q}_{t}$ are used to get the ``intermediate" hidden state:
\begin{eqnarray}
\tilde{\mathbf{s}}_{t}^{r}=\mathbf{GRU}(\mathbf{q}_{t}, \tilde{\mathbf{c}}_{t}^{r})
\end{eqnarray}
Finally, we use the ``intermediate" hidden state to update $\mathbf{K}_{(t)}^{(r-1)}$ as recording the ``intermediate"  attention status, to finish a round of operations,
\begin{eqnarray}
\mathbf{K}_{(t)}^{(r)}=\mathbf{Update}(\tilde{\mathbf{s}}_{t}^{r}, \mathbf{K}_{(t)}^{(r-1)}) \label {write}
\end{eqnarray}
After the last round ($R$) of the operations, we use $\{\tilde{\mathbf{s}}_{t}^{R}, \tilde{\mathbf{c}}_{t}^{R}, \tilde{\mathbf{a}}_{t}^{R}\}$ as the resulting states to compute a final prediction via Eq.~\ref{predict}.
Then the \textsc{Key-Memory} $\mathbf{K}_{(t)}^{(R)}$ will be transited to the next decoding step $t$+$1$, being $\mathbf{K}_{(t+1)}^{(0)}$. The details of $\mathbf{Address}$, $\mathbf{Read}$ and $\mathbf{Update}$ operations will be described later in next section.

As seen, the \textsc{KVMemAtt} mechanism can update the \textsc{Key-Memory} along with the update-chain of the decoder state to keep track of attention status and also maintains a fixed \textsc{Value-Memory} to store the representation of source sentence.  At each decoding step, the~\textsc{KVMemAtt} generates the context representation of source sentence via nontrivial transformations between the \textsc{Key-Memory} and the \textsc{Value-Memory}, and records the attention status via interactions between the two memories.
This structure allows the model to leverage possibly complex transformations and interactions between two memories, and lets the decoder choose more appropriate source context for the word prediction at each step.
Clearly, \textsc{KVMemAtt} can subsume the coverage models~\cite{Tu2016,MiEtAl2016} and the interactive attention model~\cite{mengEtAlCOLING2016} as special cases, while more generic and powerful, as empirically verified in the experiment section.

\subsection{Memory Access Operations} \label{rw_ia}
In this section, we will detail the memory access operations from round $r$-$1$ to round $r$ at decoding time step $t$. \vspace{-10pt}

\paragraph{Key-Memory Addressing} Formally, $\mathbf{K}_{(t)}^{(r)} \in \mathbb{R}^{n \times d}$ is the \textsc{Key-Memory} in round $r$ at decoding time step $t$ before the decoder RNN state update, where $n$ is the number of memory slots and $d$ is the dimension of vector in each slot. The addressed attention vector is given by
\begin{eqnarray}
\tilde{\mathbf{a}}_{t}^{r} = \mathbf{\mathbf{Address}}(\mathbf{q}_{t}, \mathbf{K}_{(t)}^{(r-1)}) \label{address_r}
\end{eqnarray}
where $\tilde{\mathbf{a}}_{t}^{r} \in \mathbb{R}^{n}$ specifies the normalized weights assigned to the slots in $\mathbf{K}_{(t)}^{(r-1)}$, with the $j$-th slot being $\mathbf{k}_{(t,j)}^{(r-1)}$. We can use content-based addressing to determine $\tilde{\mathbf{a}}_{t}^{r}$ as described in~\cite{ntmgraves2014neural} or (quite similarly) use the soft-alignment as in Eq.~\ref{match_score}. In this paper, for convenience we adopt the latter one. And the $j$-th cell of $\tilde{\mathbf{a}}_{t}^{r}$ is
\begin{eqnarray}
\tilde{\mathbf{a}}_{t,j}^{r} = \frac{exp(\tilde{e}_{t,j}^{r})}{\sum_{i=1}^{n}{exp{(\tilde{e}_{t,i}^{r})}}}
\end{eqnarray}
where $\tilde{e}_{t,j}^{r}={\mathbf{v}_a^{r}}^Ttanh(\mathbf{W}_a^{r} \mathbf{q}_{t} + \mathbf{U}_a^{r}  \mathbf{k}_{(t,j)}^{(r-1)})$.
\vspace{-10pt}

\paragraph{Value-Memory Reading} Formally, $\mathbf{V} \in \mathbb{R}^{n \times d}$ is the \textsc{Value-Memory}, where $n$ is the number of memory slots and $d$ is the dimension of vector in each slot. Before the decoder state update at time $t$, the output of reading at round $r$ is $\tilde{\mathbf{c}}_t^r$ given by
\begin{eqnarray}
\tilde{\mathbf{c}}_{t}^{r}=\sum_{j=1}^{n}\tilde{\mathbf{a}}_{t,j}^{r}\mathbf{v}_j
\end{eqnarray}
where $\tilde{\mathbf{a}}_{t}^{r} \in \mathbb{R}^{n}$ specifies the normalized weights assigned to the slots in $\mathbf{V}$.

\vspace{-5pt}
\paragraph{Key-Memory Updating} Inspired by the attentive writing operation of neural turing machines~\cite{ntmgraves2014neural}, we define two types of operation for updating the \textsc{Key-Memory}: \textsc{Forget} and \textsc{Add}.

\textsc{Forget} determines the content to be removed from memory slots. More specifically, the vector $\mathbf{F}_t^r \in \mathbb{R}^{d}$ specifies the values to be forgotten or removed on each dimension in memory slots, which is then assigned to each slot through normalized weights $\mathbf{w}_{t}^r$. Formally, the memory (``intermediate") after \textsc{Forget} operation is given by
\begin{eqnarray}
\tilde{\mathbf{k}}^{(r)}_{t,i} =\mathbf{k}^{(r-1)}_{t,i}(1-\mathbf{w}_{t, i}^r\cdot \mathbf{F}_t^{r}),  \hspace{10pt} i = 1, 2, \cdots, n
\end{eqnarray}
where
\begin{itemize}
\item $\mathbf{F}_t^r=\sigma(\mathbf{W}_F^r, \tilde{\mathbf{s}}_{t}^{r})$ is parameterized with $\mathbf{W}_F^r \in \mathbb{R}^{d \times d}$, and $\sigma$ stands for the $Sigmoid$ activation function, and $\mathbf{F}_t^r \in \mathbb{R}^{d}$;
\item $\mathbf{w}_t^r \in \mathbb{R}^{n}$ specifies the normalized weights assigned to the slots in $\mathbf{K}_{(t)}^{(r)}$, and $\mathbf{w}_{t,i}^r$ specifies the weight associated with the $i$-th slot. $\mathbf{w}_t^r$ is determined by
\begin{eqnarray}
\mathbf{w}_{t}^{r}=\mathbf{Address}(\tilde{\mathbf{s}}_{t}^{r}, \mathbf{K}_{(t)}^{(r-1)})
\end{eqnarray}
\end{itemize}

\textsc{Add} decides how much current information should be written to the memory as the added content
\begin{eqnarray}
\mathbf{k}^{(r)}_{t,i}=\tilde{\mathbf{k}}^{(r)}_{t,i} + \mathbf{w}_{t,i}^r\cdot \mathbf{A}_t^r,  \hspace{20pt} i = 1, 2, \cdots, n
\end{eqnarray}
where $\mathbf{A}_t^r=\sigma(\mathbf{W}_A^r,  \tilde{\mathbf{s}}_{t}^{r})$ is parameterized with $\mathbf{W}_A^r \in \mathbb{R}^{d \times d}$, and $\mathbf{A}_t^r \in \mathbb{R}^{d}$.
Clearly, with \textsc{Forget} and \textsc{Add} operations, \textsc{KVMemAtt} potentially can modify and add to the \textsc{Key-Memory} more than just history of attention.
\vspace{0pt}

\subsection{Training}

\paragraph{EOS-attention Objective:}
The translation process is finished when the decoder generates $eos_{trg}$ (a special token that stands for the end of target sentence). Therefore, accurately generating $eos_{trg}$ is crucial for producing correct translation. Intuitively, accurate attention for the last word (i.e. $eos_{src}$) of source sentence will help the decoder accurately predict $eos_{trg}$. When to predict $eos_{trg}$ (i.e. $y_m$), the decoder should highly focus on $eos_{src}$ (i.e. $x_n$). That is to say the attention probability of $a_{m,n}$ should be closed to 1.0. And when to generate other target words, such as $y_1,y_2,\cdots, y_{m-1}$, the decoder should not focus on $eos_{src}$ too much. Therefore we define

\begin{eqnarray}
\mathbf{\textsc{Atteos}}=\sum_{t=1}^{m}{Att_{eos}(t,n)} \label{eos_att_obj}
\end{eqnarray}
where
\begin{align}
Att_{eos}(t,n)=\left\{
\begin{aligned}
& a_{t, n},          & t<m \\
& 1.0 - a_{t,n},  & t =m \\
\end{aligned}
\right.
\end{align}

\paragraph{NMT Objective:}
We plag Eq.~\ref{eos_att_obj} to Eq.~\ref{nmt_obj}, we have
\begin{align}
\mathbf{\theta^{*}} & = &  \arg\max_{\theta}\sum_{s=1}^{S} & \Big( \log p(\mathbf{y}^{s}|\mathbf{x}^{s})  - \lambda * \mathbf{\textsc{Atteos}^s} \Big) \label{obj_final}
\end{align}
The EOS-attention objective can assist the learning of \textsc{KVMemAtt} and guiding the parameter training, which will be verified in the experiment section.

\section{Experiments}\label{experiments}

\begin{table*}[t!]
\centering
\renewcommand{\arraystretch}{1.1}
\scalebox{0.99}{
\begin{tabular}{l|l||c c||cccc|c}
\bf \textsc{System}	& \bf \textsc{Architecture}	&\bf \# Para. & \bf Speed & \bf \textsc{MT03} &	\bf \textsc{MT04}	& \bf \textsc{MT05}	& \bf \textsc{MT06}	& \bf \textsc{Ave.}\\
\hline
\hline
\multicolumn{9}{c}{\em Existing end-to-end NMT systems} \\
\hline
\cite{Tu2016}				& \textsc{Coverage}   &  --  &	-- & 33.69	& 38.05	& 35.01	 & 34.83	& 35.40\\
\cite{mengEtAlCOLING2016} 	& \textsc{MemAtt}		&  --  &	-- & 35.69	& 39.24	& 35.74	 & 35.10	& 36.44\\
\cite{WangLLL16}  			& \textsc{MemDec}		&  --  &	-- & 36.16	& 39.81	& 35.91	& 35.98	& 36.95\\
\cite{zhangEtalACL2017} 	& \textsc{Distortion} &  --  &	-- & 37.93	& 40.40	& 36.81	&35.77	& 37.73\\	
\hline
\hline
\multicolumn{9}{c}{\em Our end-to-end NMT systems} \\
\hline
\multirow{6}{*}{this work}  			&   \textsc{RNNSearch}		& 53.98M  & 2773  & 36.02     & 38.65     & 35.64     & 35.80  & 36.53\\
    &   ~~~~~+ \textsc{MemAtt}   & 55.03M  & 2319  & 36.93*+  & 40.06*+  & 36.81*+ & 37.39*+  & 37.80  \\
\cline{2-9}
& ~~~~~+ \textsc{KVMemAtt-R1} & 55.03M  & 2263	& 37.56*+    & 40.56*+    & 37.83*+    & 37.84*+  & 38.45\\
& ~~~~~~~~~~\textsc{+AttEosObj}	&   55.03M   & 1992  & 37.87*+     &40.64+ 	  & 38.01+      & 38.13*+    & 38.66\\
\cdashline{2-9}
&   ~~~~~+ \textsc{KVMemAtt-R2}		& 58.18M  & 1804  & 38.08+    & 40.73+    & 38.09+    & 38.80*+  & 38.93\\
&   ~~~~~~~~~~+ \textsc{AttEosObj}	& 58.18M  & 1676  & 38.40*+    & 41.10*+    & 38.73*+    & 39.08*+  & 39.33\\
\end{tabular}
}
\caption{\label{t:main-result} Case-insensitive BLEU scores (\%) on Chinese$\Rightarrow$English translation task. ``Speed'' denotes the training speed (words/second).
``R1'' and ``R2'' denotes \textsc{KVMemAtt} with one and two rounds, respectively.
``$\textsc{+AttEosObj}$" stands for adding the EOS-attention objective. The “*” and ``+" indicate that the results are significantly (p$<$0.01 with \emph{sign-test}) better than the adjacent above system and the ``$\textsc{RNNSearch}$".
} \vspace{-5pt}
\end{table*}

\subsection{Setup}

We carry out experiments on Chinese$\Rightarrow$English (Zh$\Rightarrow$En) and German$\Leftrightarrow$English (De$\Leftrightarrow$En) translation tasks. For Zh$\Rightarrow$En, the training data consist of 1.25M sentence pairs extracted from LDC corpora. 
We choose NIST 2002 (MT02) dataset as our valid set, and NIST 2003-2006 (MT03-06) datasets as our test sets. For De$\Leftrightarrow$En, we perform our experiments on the corpus provided by WMT17, which contains 5.6M sentence pairs. We use newstest2016 as the development set, and newstest2017 as the testset. 
We measure the translation quality with BLEU scores~\cite{papineni2002bleu}.\footnote{For Zh$\Rightarrow$EEn task, we apply case-insensitive NIST BLEU \emph{mteval-v11b.pl}. For De$\Leftrightarrow$En tasks, we tokenized the reference and evaluated the performance with case-sensitive \emph{multi-bleu.pl}. The metrics are exactly the same as in previous work.} 

In training the neural networks, we limit the source and target vocabulary to the most frequent 30K words for both sides in Zh$\Rightarrow$En task, covering approximately 97.7\% and 99.3\% of two corpus respectively.  For De$\Leftrightarrow$En, sentences are encoded using byte-pair encoding~\cite{sennrichACL2016}, which has a shared source-target vocabulary of about 36000 tokens. 
The parameters are updated by SGD and mini-batch (size 80) with learning rate controlled by AdaDelta~\cite{adadelta} ($\epsilon=1e^{-6}$ and $\rho=0.95$). 
We limit the length of sentences in training to 80 words for Zh$\Rightarrow$En and 100 sub-words for De$\Leftrightarrow$En. The dimension of word embedding and hidden layer is 512, and the beam size in testing is 10. We apply dropout on the output layer to avoid over-fitting~\cite{hinton2012improving}, with dropout rate being 0.5.
Hyper parameter $\lambda$ in Eq.~\ref{obj_final} is set to 1.0.
The parameters of our \textsc{KVMemAtt}  (i.e., encoder and decoder, except for those related to \textsc{KVMemAtt}) are initialized by the pre-trained baseline model. 

\subsection{Results on Chinese-English}

We compare our \textsc{KVMemAtt} with two strong baselines: 
1) \textsc{RNNSearch}, which is our in-house implementation of the attention-based NMT as described in Section~\ref{back}; 
and 2) \textsc{RNNSearch+MemAtt}, which is our implementation of interactive attention~\cite{mengEtAlCOLING2016} on top of our \textsc{RNNSearch}. 
Table~\ref{t:main-result} shows the translation performance.

\vspace{-10pt}
\paragraph{Model Complexity}

\textsc{KVMemAtt} brings in little parameter increase. Compared with \textsc{RNNsearch}, \textsc{KVMemAtt-R1} and \textsc{KVMemAtt-R2} only bring in 1.95\% and 7.78\% parameter increase. Additionally, introducing \textsc{KVMemAtt} slows down the training speed to a certain extent (18.39\%$\sim$39.56\%). When running on a single GPU device Tesla P40, the speed of the \textsc{RNNsearch} model is 2773 target words per second, while the speed of the proposed models is 1676$\sim$2263 target words per second.

\begin{table}[t]
\renewcommand{\arraystretch}{1.1}
\scalebox{0.99}{
\centering
\begin{tabular}{l|cc|cc}
\bf \textsc{Architecture} & \bf \textsc{BLEU} & \bf $\bigtriangleup$ & \bf \textsc{AER}  & \bf $\bigtriangledown$ \\
\hline\hline
\textsc{RNNsearch}           	& 24.65   & - 		& 39.97 & - \\
~~~+ \textsc{MemAtt}		& 25.36   & +0.71	& 37.98 & -1.99 \\
\hline
~~~+ \textsc{KVMemAtt-R1}		& 25.55   & +0.90	& 37.70 & -2.27 \\
~~~~~~~~+ \textsc{AttEosObj}   	& 25.93   & +1.28	& 36.89 & -3.08 \\
\hdashline
~~~+ \textsc{KVMemAtt-R2}		& 26.43   & +1.78	& 35.70 & -4.27 \\
~~~~~~~~+ \textsc{AttEosObj}   	& 27.19   & +2.54	& 35.39 & -4.58 \\
\end{tabular}
}
\caption{\label{t:align-result} Performance on manually aligned Chinese-English test set. Higher BLEU score and lower AER score indicate better quality.} \vspace{-10pt}
\end{table}

\begin{figure}[t]
\begin{center}
      \includegraphics[width=0.27\textwidth]{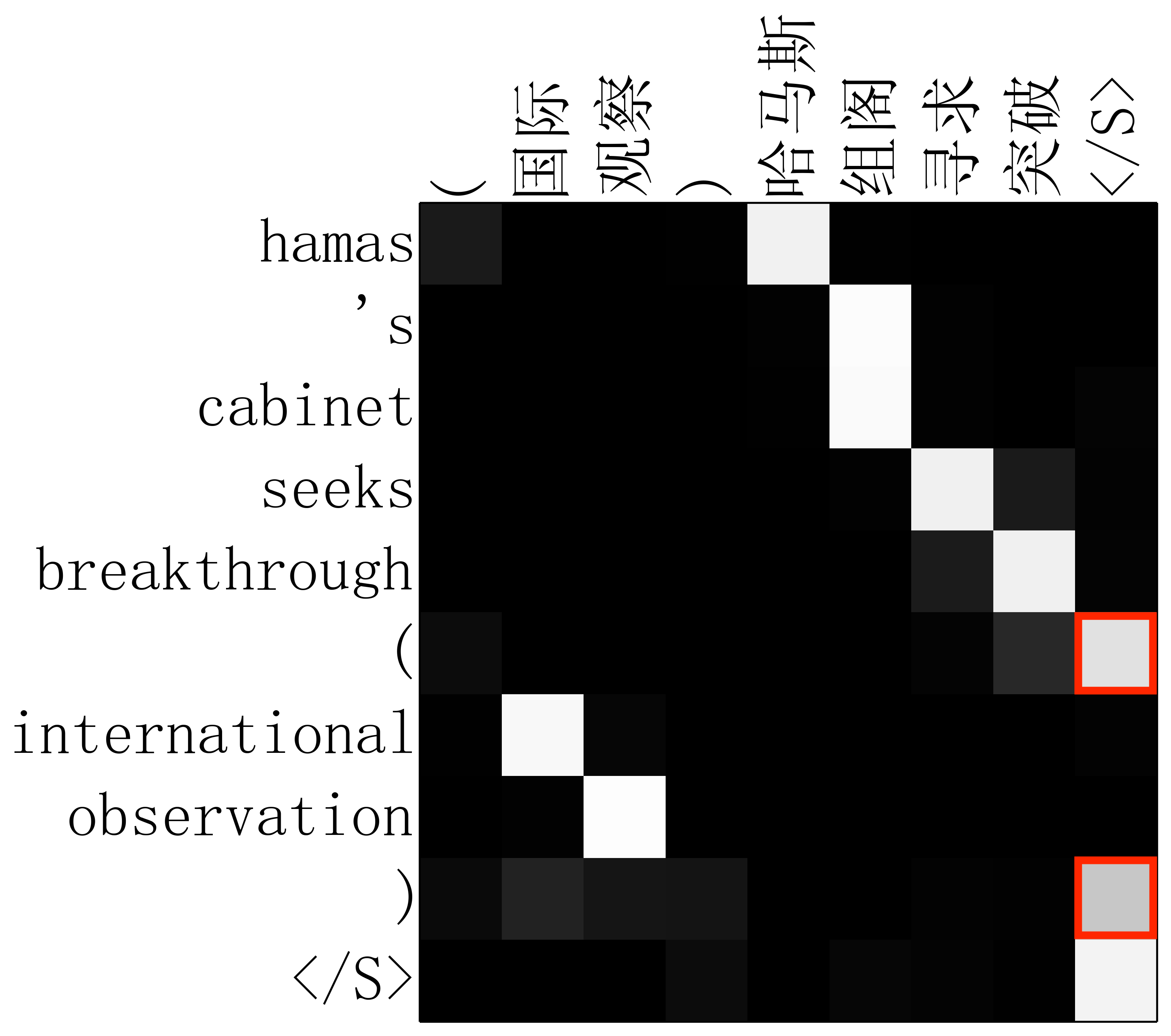} \hfill
      \includegraphics[width=0.172\textwidth]{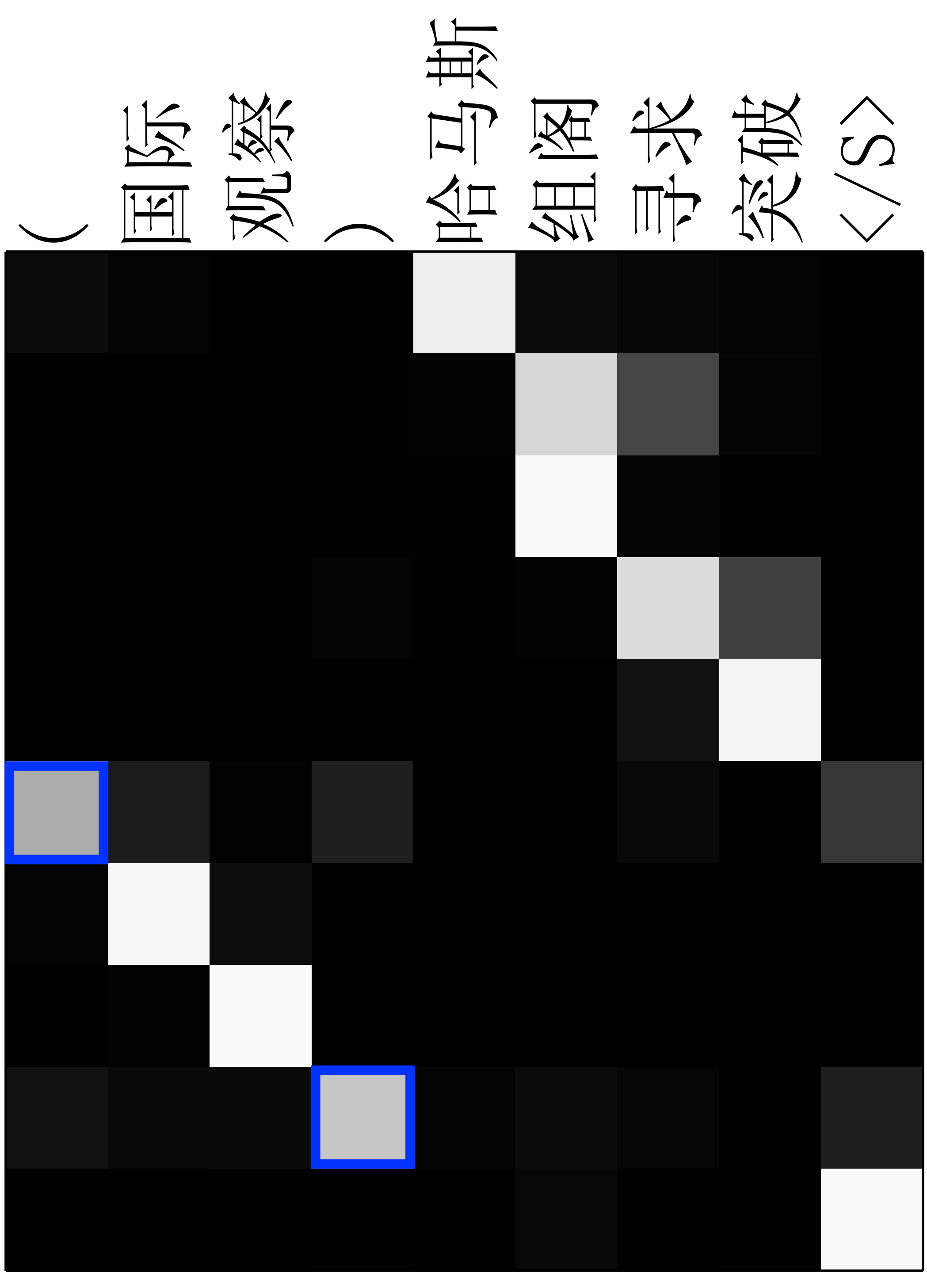}
       \caption{Example attention matrices of \textsc{KVMemAtt-R2} generated from the first round (left) and the second round (right). We highlight some bad links with read frame and good links with blue frame.}  \label{example1}
\end{center}  \vspace{-10pt}
\end{figure}

\vspace{-10pt}
\paragraph{Translation Quality}

\textsc{KVMemAtt} with one round memory access (\textsc{KVMemAtt-R1}) achieves significant improvements over \textsc{RNNsearch} by 1.92 BLEU and over \textsc{RNNsearch+MemAtt} by 0.65 BLEU averagely on four test sets. It indicates that our key-value memory-augmented attention mechanism can give more effective power for the attention via nontrivial transformations and interactions between the \textsc{Key-Memory} and the \textsc{Value-Memory}. 
The two rounds counterpart (\textsc{KVMemAtt-R2}) can further improve the performance by 0.48 BLEU on average. It confirms our hypothesis that the decoder can benefit from the re-attention process, which considers the ``intermediate" attention result achieved in the early stage and then makes a more accurate decision. 
However, we find that adding more than two rounds of memory access operations into \textsc{KVMemAtt} does not lead to better translation performance (not shown in the table). One possible reason is that memory access with more rounds leads to more updating operations (i.e. attentive writing) on the \textsc{Key-Memory}, which may be difficult to optimize within our current architecture design. We will leave it as future work.

The contribution of adding EOS-attention Objective is to assist the learning of attention, and guiding the parameter training.
It consistently improves translation performance over different variants of \textsc{KVMemAtt}.
It gives a further improvement of 0.40 BLEU points over \textsc{KVMemAtt-R2}, which is 2.80 BLEU points better than \textsc{RNNSearch}.

\vspace{-10pt}
\paragraph{Alignment Quality}
Intuitively, our~\textsc{KVMemAtt} can enhance the attention and therefore improve the word alignment quality. To confirm our hypothesis, we carry out experiments of the alignment task on the evaluation dataset from~\cite{Liuyang2015}, which contains 900 manually aligned Chinese-English sentence pairs. We use the alignment error rate (AER)~\cite{och2003systematic} as the evaluation metric for the alignment task.
Table~\ref{t:align-result} lists the BLEU and AER scores. As expected, our~\textsc{KVMemAtt} achieves better BLEU and AER scores (the lower the AER score, the better the alignment quality) than the strong baseline systems. Additionally, the results also indicate that the EOS-attention objective can assist the learning of attention-based NMT, since adding this objective yields better alignment performance. 
By visualizing the attention matrices, we found that the attention qualities are improved from the first round to the second round as expected, as shown in Figure~\ref{example1}.

\begin{table}[tb]
\begin{center}
\scalebox{0.99}{
\begin{tabular}{l|c|c}
\textsc{Systems}  & \textsc{UnderTran} & \textsc{OverTran}  \\
\hline\hline
\textsc{RNNSearch}           & 13.1\%  & 2.7\%       \\
~~~+ \textsc{KVMemAtt-R2}     & 9.7\%    & 1.3\%      \\
\end{tabular}
}
\end{center} 
\caption{\label{t:adequacy-result} Subjective evaluation of translation adequacy. Numbers denote percentages of source words.}
\end{table}

\begin{table*}[t!]
\centering
\renewcommand{\arraystretch}{1.1}
\scalebox{0.99}{
\begin{tabular}{l|l|c|c}
\bf \textsc{System}  & \bf \textsc{Architecture}  & \bf \textsc{De$\Rightarrow$En} & \bf \textsc{En$\Rightarrow$De}\\
\hline
\hline
\multicolumn{4}{c}{\em Existing end-to-end NMT systems} \\
\hline
\cite{RiktersetalWMT2017}  & cGRU + dropout + named entity forcing + \emph{synthetic data}	& 29.00 & 22.70\\
\cite{escolanoWMT2017}   & Char2Char + rescoring with inverse model + \emph{synthetic data}	& 28.10 & 21.20\\
\cite{UENSfW2017} & cGRU +  \emph{synthetic data} &  32.00 & 26.10\\
\hline
\cite{Tu2016}             & \textsc{RNNSearch + Coverage} & 28.70 & 23.60 \\
\cite{Zheng:2018:TACL}	     & \textsc{RNNSearch + Past-Future-Layers} & 29.70 & 24.30\\
\hline
\hline
\multicolumn{4}{c}{\em Our end-to-end NMT systems} \\
\hline
\multirow{2}{*}{this work} 	& \textsc{RNNsearch}  & 29.33   & 23.83\\
& ~~~~~+ \textsc{MemAtt}  & 30.13   & 24.62 \\
\cdashline{2-4}
& ~~~~~+ \textsc{KVMemAtt-R2 + AttEosObj}  & 30.98   & 25.39 \\
\end{tabular}
}
\caption{\label{t:ende-result} Case-sensitive BLEU scores (\%) on German$\Leftrightarrow$English translation task. ``\emph{synthetic data}” denotes additional 10M monolingual sentences, which is not used in this work.} \vspace{-10pt}
\end{table*}

\vspace{-10pt}
\paragraph{Subjective Evaluation}

We did a subjective evaluation to investigate the benefit of incorporating \textsc{KVMemAtt} to NMT, especially on alleviating the issues of over- and under-translations. Table~\ref{t:adequacy-result} lists the  translation adequacy of the \textsc{RNNSearch} baseline and our $\textsc{KVMemAtt-R2}$ on the randomly selected 100 sentences from test sets. From Table~\ref{t:adequacy-result} we can see that,  compared with the baseline system, our approach can decline the percentages of source words which are under-translated from 13.1\% to 9.7\% and which are over-translated from 2.7\% to 1.3\%. The main reason is that our \textsc{KVMemAtt} can keep track of attention status and generate more appropriate source context for predicting the next target word at each decoding step. 

\subsection{Results on German-English}
We also evaluate our model on the WMT17 benchmarks on bidirectional German$\Leftrightarrow$English translation tasks, as listed in Table~\ref{t:ende-result}. Our baseline achieves even higher BLEU scores to the state-of-the-art NMT systems of WMT17, which do not use additional synthetic data.\footnote{\cite{UENSfW2017} obtains better BLEU scores than our model, since they use large scaled synthetic data (about 10M). It maybe unfair to compare our model to theirs directly. }
Our proposed model consistently outperforms two strong baselines (i.e., standard and memory-augmented attention models) on both De$\Rightarrow$En and En$\Rightarrow$De translation tasks.
These results demonstrate that our model works well across different language pairs.

\section{Related Work}
Our work is inspired by the key-value memory networks~\cite{millerEMNLP2016} originally proposed for the question answering, and has been successfully applied to machine translation~\cite{Gu:2017:arXiv,gehring2017convs2s,VaswaniEtal2017,Tu:2018:TACL}.
In these works, both the key-memory and value-memory are fixed during translation. Different from these works, we update the \textsc{Key-Memory} along with the update-chain of the decoder state via attentive writing operations (e.g. \textsc{Forget} and \textsc{Add}).

Our work is related to recent studies that focus on designing better attention models~\cite{luongEMNLP2015,cohnEtAl2016,FengLLZ16,Tu2016,MiEtAl2016,zhangEtalACL2017}.~\cite{luongEMNLP2015} proposed to use a global attention to attend to all source words and a local attention model to look at a subset of source words.~\cite{cohnEtAl2016} extended the attention-based NMT to include structural biases from word-based alignment models.~\cite{FengLLZ16} added implicit distortion and fertility models to attention-based NMT.~\cite{zhangEtalACL2017} incorporated distortion knowledge into the attention-based NMT.~\cite{Tu2016,MiEtAl2016} proposed coverage mechanisms to encourage the decoder to consider more untranslated source words during translation. These works are different from our~\textsc{KVMemAtt}, since we use a rather generic key-value memory-augmented framework with memory access (i.e. \emph{address}, \emph{read} and \emph{update}). 

Our work is also related to recent efforts on attaching a memory to neural networks 
\cite{ntmgraves2014neural} and exploiting memory~\cite{TangMLLY16,WangLLL16,feng2017memory,mengEtAlCOLING2016,Wangxing2017} during translation.~\cite{TangMLLY16} exploited a phrase memory stored in symbolic form for NMT.~\cite{WangLLL16} extended the NMT decoder by maintaining an external memory, which is operated by reading and writing operations.~\cite{feng2017memory} proposed a neural-symbolic architecture, which exploits a memory to provide knowledge for infrequently used words.
Our work differ at that we augment attention with a specially designed interactive key-value memory, which allows the model to leverage possibly complex transformations and interactions between the two memories via single- or multi-rounds of memory access in each decoding step.  

\vspace{-3pt}
\section{Conclusion and Future Work}
\vspace{-2pt}
We propose an effective \textsc{KVMemAtt} model for NMT, which maintains a timely updated key-memory to track attention history and a fixed value-memory to store the representation of source sentence during translation. Via nontrivial transformations and iterative interactions between the two memories, our \textsc{KVMemAtt} can focus on more appropriate source context for predicting the next target word at each decoding step. Additionally, to further enhance the attention, we propose a simple yet effective attention-oriented objective in a weakly supervised manner.
Our empirical study on Chinese$\Rightarrow$English, German$\Rightarrow$English and English$\Rightarrow$German translation tasks shows that \textsc{KVMemAtt} can significantly improve the performance of NMT. 

For future work, we will consider to explore more rounds of memory access with more powerful operations on key-value memories to further enhance the attention. Another interesting direction is to apply the proposed approach to Transformer~\cite{VaswaniEtal2017}, in which the attention model plays a more important role.

\bibliographystyle{named}
\bibliography{ijcai18}

\end{document}